\pdfoutput=1
\documentclass[11pt]{article}
\usepackage[final]{emnlp2021}  % 在这里填final还是review，对应sty里面的finalcopy选项

\usepackage{times}
\usepackage{latexsym}

\usepackage[T1]{fontenc}
\usepackage[utf8]{inputenc}

\usepackage{microtype}

\usepackage{CJKutf8}
\usepackage{tabularx}
\usepackage{pifont}% http://ctan.org/pkg/pifont
\usepackage{graphicx}
\usepackage{romannum}
\usepackage{amsmath}
\usepackage{booktabs} % for tabel
\usepackage{fixltx2e} % for text subscript
\usepackage{multirow}
\title{What Have Been Learned \& What Should Be Learned? \\ An Empirical Study of How to Selectively Augment Text for Classification}

\author{Biyang Guo\thanks{\ \ All authors contribute equally.}, \ Sonqiao Han\footnotemark[1], \ Hailiang Huang\footnotemark[1] \ \thanks{\ \ Corresponding author.}\\
        AI Lab, School of Information Management and Engineering, \\Shanghai University of Finance and Economics,
    Shanghai, China, 200433 \\}

\begin{document}
\maketitle
\begin{abstract}
Text augmentation techniques are widely used in text classification problems to improve the performance of classifiers, especially in low-resource scenarios. Whilst lots of creative text augmentation methods have been designed, they augment the text in a \textit{non-selective} manner, which means the less important or noisy words have the same chances to be augmented as the informative words, and thereby limits the performance of augmentation. In this work, we systematically summarize three kinds of \textit{role keywords}, which have different functions for text classification, and design effective methods to extract them from the text. Based on these extracted \textit{role keywords}, we propose \textbf{STA} (Selective Text Augmentation) to selectively augment the text, where the informative, class-indicating words are emphasized but the irrelevant or noisy words are diminished. Extensive experiments on four English and Chinese text classification benchmark datasets demonstrate that \textbf{STA} can substantially outperform the non-selective text augmentation methods. 
\end{abstract}

\begin{figure*}[t]
 \centering
\includegraphics[width=\textwidth]{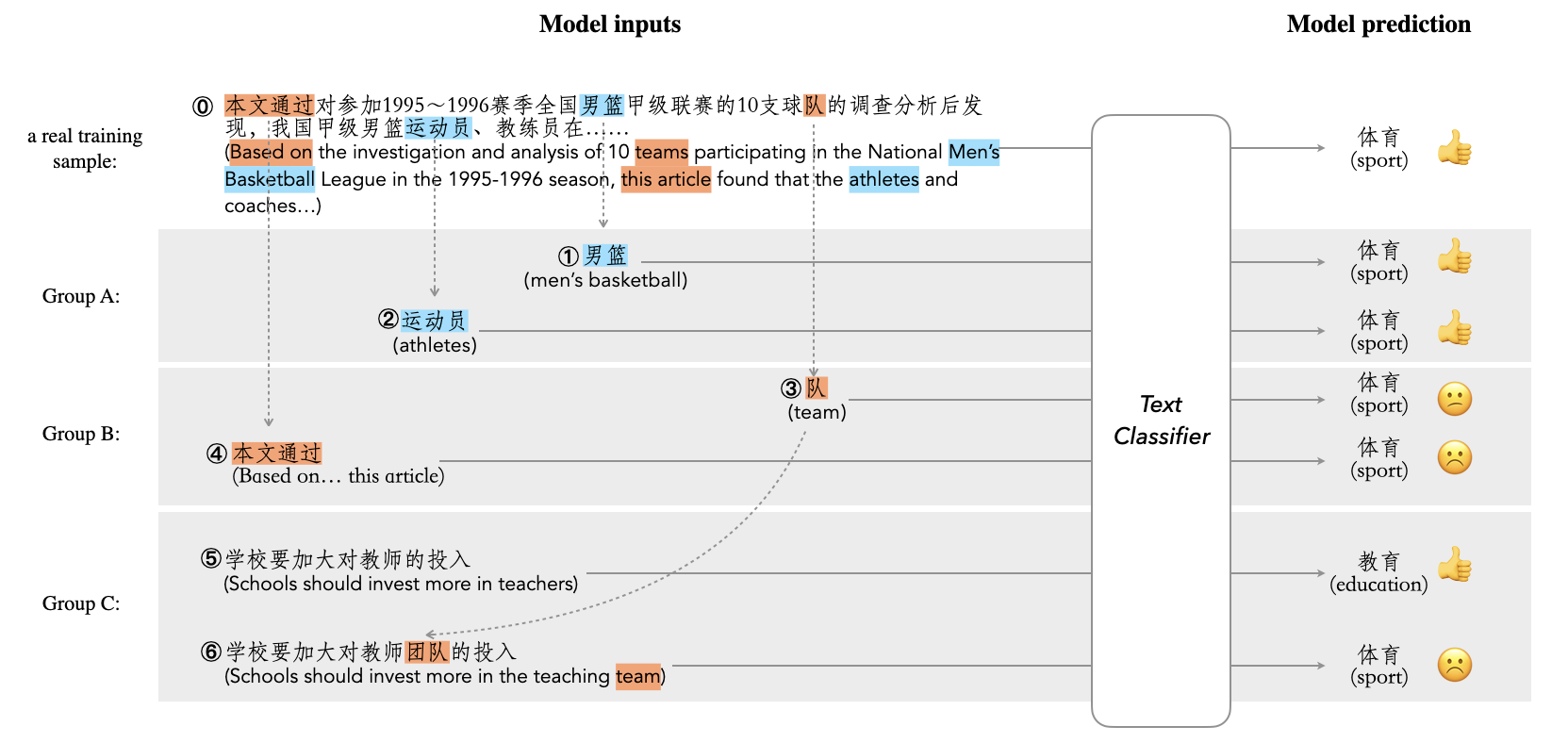} % Reduce the figure size so that it is slightly narrower than the column.
\caption{A case study to understand what have been learned by a text classifier. The text classifier is trained on the FDUCNews dataset. Sentence 0 is sampled from the dataset with class "sport". Group A shows that sport-related words are correctly learned as strong features of the "sport" . Group B illustrates that some words irrelevant to "sport" are also learned as features of "sport". Group C subsequently shows that these fake features may mislead the prediction.}
\label{case_study}
\end{figure*}

\section{Introduction}
Text classification is one of the fundamental tasks in Natural Language Processing (NLP), which has wide application in news filtering, paper categorization, sentiment analysis and so on. Plenty of algorithms, especially deep learning models have achieved great success in text classification, such as recurrent neural networks (RNN) \cite{liu2016recurrent, wang2018topic}, convolutional networks (CNN) \cite{kim2014convolutional} and BERT \cite{devlin2018bert}. The success of deep learning is usually built on the large training data with good quality, which is often difficult to obtain in real applications. Therefore, text augmentation techniques have attracted more and more attention both in academic and industrial communities to improve the generalization ability of text classification models when training data is limited. 

Effective text augmentation techniques include text editing methods \cite{sr2011,sr2015a,wei2019eda} and back-translation \cite{back_translation1,back_translation2}. Existing usages of these techniques are in a \textit{non-selective} manner: augmentation is applied to all words in the given text without difference. For example, in EDA \cite{wei2019eda}, the words to be replaced, deleted or swapped are all randomly chosen from the text. However, different words in the text have different importance to the classification task. If the less important words are chosen to be augmented, the augmentation will have little performance gain. If the noisy or misleading words are augmented, the augmentation will even hurt the performance. 

By intuition, by carefully selecting the most informative parts to be augmented and avoiding the influence of noisy words, the text augmentation methods are likely to achieve a higher classification accuracy.

In this work, we first explore \textit{what have been learned} by text classifiers through analysing the roles of different words in text classification tasks. We find that what text classifiers learned from the data aren't always \textit{what should be learned} to correctly solve the classification problems. Based on these observations, we  summarize three types of \textit{role keywords}: class-indicating words, fake class-indicating words and class-irrelevant words, each of which has different effect on the classification task. Then we introduce effective methods to extract these \textit{role keywords} from the given text in order to assist our text augmentation. After that, we present \textbf{STA} (\textbf{S}elective \textbf{T}ext \textbf{A}ugmentation) to selectively augment the words in the text according to their different roles. Extensive experiments conducted on four benchmark text classification datasets reveal that \textbf{STA} outperforms traditional non-selective text augmentation baselines in a large margin. We conclude our contributions as follows:
\begin{itemize}
    \item We present three kinds of \textit{role keywords} with different influences for text classification, and introduce effective methods to extract them from the text, which can be used to boost text augmentation or other down-stream tasks.
    \item We propose \textbf{STA} to selectively augment the text based on the extracted \textit{role keywords} and conduct substantial experiments to prove its advantage over non-selective text augmentation methods.
\end{itemize}

\section{What Have Been Learned \& What Should Be Learned by A Text Classifier?}
 In this section, we will first look into a real text classification case to see what is actually learned by a model. Based on the observations from this case study, we will discuss how to separate the words from a piece of text according to their different roles for text classification, and what should or shouldn't be learned by a classifier for better generalization ability.

\subsection{A Case Study to Understand What Is Learned by Classifiers}
To explore what is learned by a text classifier, we trained a BERT-based model on FDUCNews dataset\footnote{http://www.nlpir.org}. To simplify the discussion, we only choose four classes from the dataset: "politics", "sport", "education" and "computer". We split the train, validate and test set by 8:2:2 and the trained model obtained a test accuracy at 98.92\% which means the model already performs quite well in this dataset. We then use some special sentences to test the model's generalization performance which are shown in Figure \ref{case_study}.

Sentence $0$ is chosen from the training samples with label "sport". The sentences in group A and B are phrases extracted from sentence $0$. From the predictions of group A, we can see that words "basketball" and "athletes" are correctly classified to "sport" class. These words are highly related to sports and learned as apparent indicators of "sport". However, phrases like "team", "based on... this article" in group B are irrelevant to the meaning of "sport" are also predicted as "sport", which shows that the model have learned some incorrect features about "sport". Group C subsequently illustrates that these incorrect features can easily mislead the prediction: simply inserting the word "team" to a education-related sentence can change the prediction from "education" to "sport". By carefully checking the training samples in the dataset, we find that the phrase "based on...this article" and word "team" are very frequent in documents belonging to class "sport" while not so common in other classes. This bias of dataset leads the model to mistakenly view words like "team" as strong features related to "sport" class, resulting in the false prediction of sentence $6$.

\begin{figure}[h]
 \centering
\includegraphics[width=0.45\textwidth]{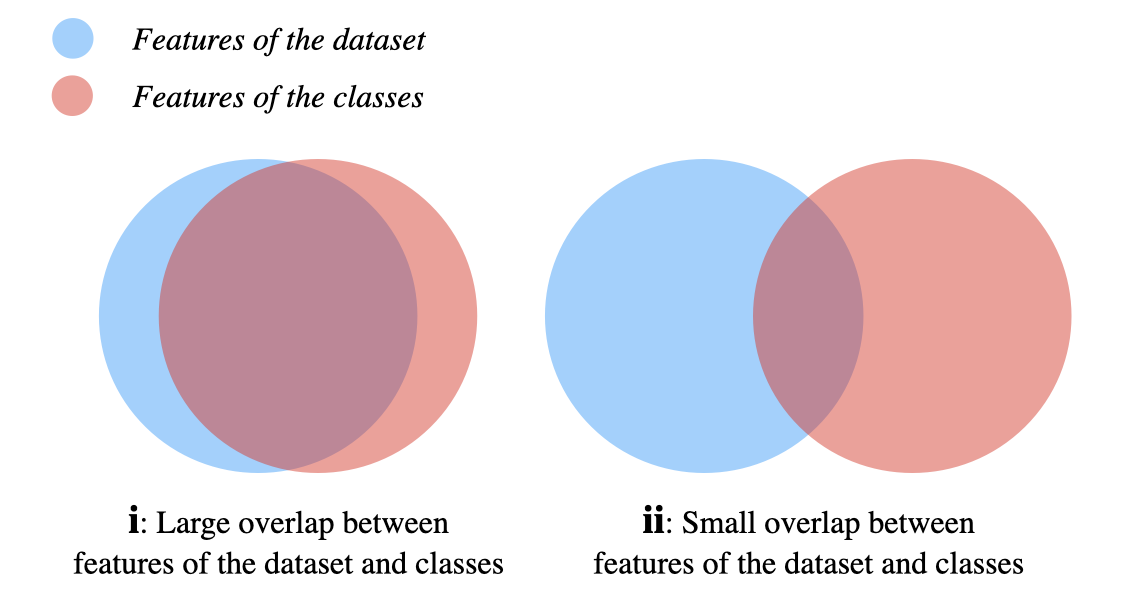} % Reduce the figure size so that it is slightly narrower than the column.
\caption{Illustration of the relation between the features of dataset and features of classes. In scenario \textbf{\romannum{1}}, the two parts share a large overlap, thus the samples from the dataset can well describe the meaning of the corresponding class. Scenario \textbf{\romannum{2}} describes a situation where the two parts have very small overlap. In this case, the samples from the dataset can poorly represent their corresponding labels.}
\label{features}
\end{figure}

The above examples are simple but thought-provoking, which reveal the fact that the classifiers may be ill-trained even if they have high accuracy in test set. Specifically, \textit{what the model actually learned are the \textbf{features of the dataset}, rather than the \textbf{features of the classes}}. This is because the model knows nothing about what class labels actually mean and what words are relevant to these labels at the beginning. The model only learn why some samples are grouped together, regardless of the meaning of the groups. That is to say, all the words that co-occur frequently with certain class but less often in other classes may become the features of this class. These features are useful for prediction within the dataset, but some of them might be irrelevant to the label itself and even harmful for generalization. An illustration of the relation between the two kinds of features are shown in Figure\ref{features}. The red circle represents the features of the classes, which are what we want to model to learn. The blue circle represents the features from the training samples, which can also be viewed as what classifiers can learn from the dataset. The area of the overlap between the two set indicates how much the training samples can represent its corresponding class. When the dataset is comprehensive and of high quality, the relation can be represented by scenario \textbf{\romannum{1}}, where the training samples and their class share lots of common features. Trained with this kind of dataset, classifier can often have good generalization ability. However, if the dataset is small and biased which can hardly represent what the corresponding classes actually are, the features of the dataset and classes will only have a small overlap, like scenario \textbf{\romannum{2}}. Using such dataset for training, the classifier can only learn a small fraction of the categories' features, while learn a lot of incorrect, dataset-specific features about the categories, resulting in unsatisfactory  generalization ability.

\subsection{Role keywords Provide Insights of How to Train A Better Classifier}
The analysis from the above case study tells us there exists a gap between what is learned by a text classifier and what we want it to learn. In low resource scenarios, the gap will be severely large and the trained model is hard to generalize to new samples. To improve the classifier's performance with limited resource, we should meticulously examine the training samples and figure out what should and shouldn't learned, what should be emphasised and what should be weakened. 

All the words in a training text sample are candidate features of its class. We can score these candidates through two different metrics: \\
$\bullet$ \textbf{Correlation} with the class. This measures how frequent a word co-occurs with a class while not with other classes in the given dataset.\\
$\bullet$ \textbf{Similarity} with the class label, which measures how semantically similar a word is with the class label.

\begin{figure}[h]
 \centering
\includegraphics[width=0.45\textwidth]{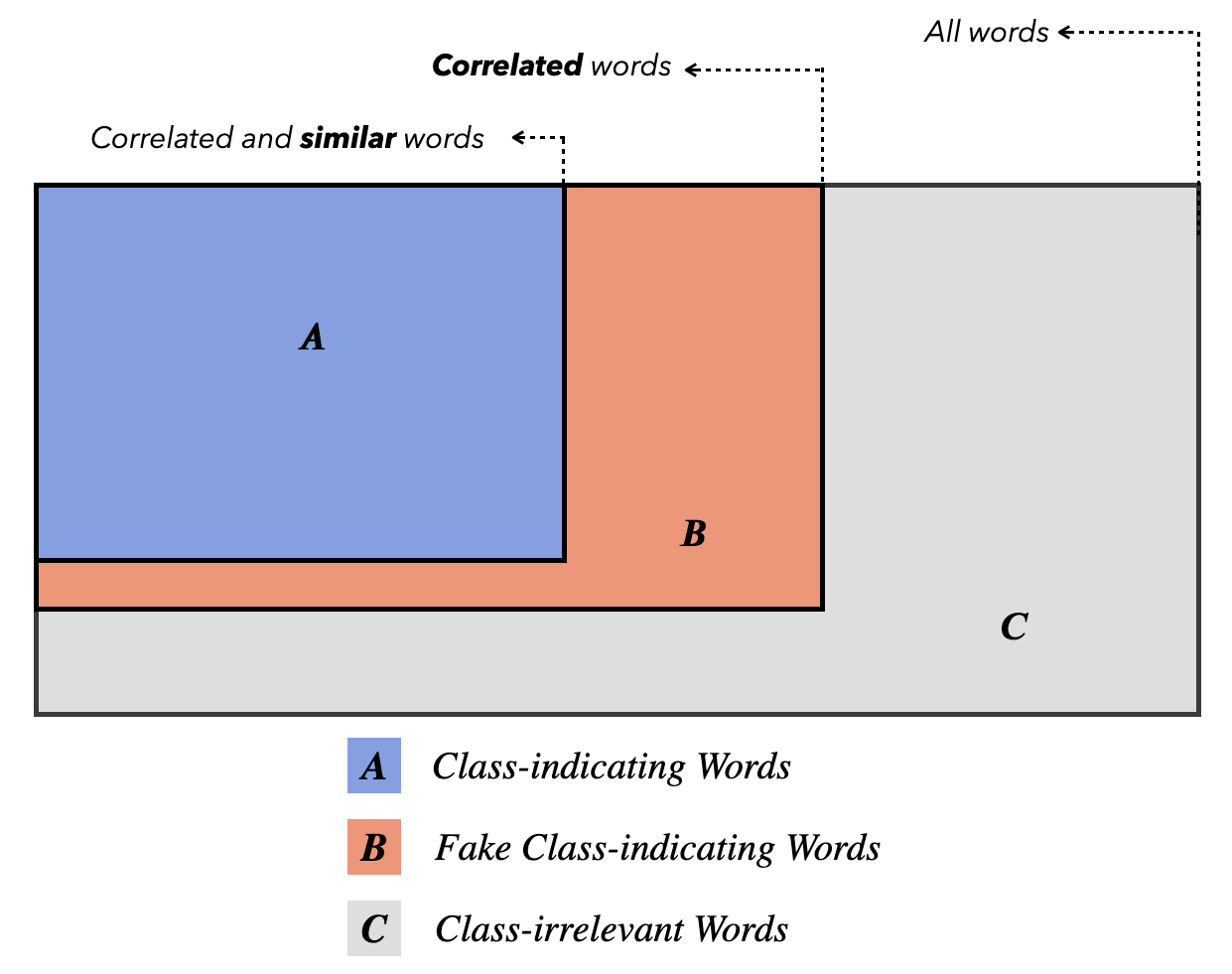} % Reduce the figure size so that it is slightly narrower than the column.
\caption{An illustration of how to obtain three kinds of \textit{role keywords} from a piece of text. The three rectangles represent "all words", "correlated words" and "correlated and similar words" respectively. These rectangles form three non-overlapping parts in different colors, which are the \textit{role keywords}.}
\label{role_kws}
\end{figure}

These metrics can be viewed as two filters and be used to divide all the words from a given text sample into three parts, as is shown in Figure\ref{role_kws}. The biggest rectangle represents all the words from the text sample. We first select out those words with high correlation level with the class, which we call "correlated words", as depicted as the second largest rectangle. We can then screen out words semantically similar with the sample's category, named as "correlated and similar words". Finally, the words from the given sample are separated into three non-overlapping types: A, B and C, which we call \textit{role keywords} in this paper. These three types of words have different effects on text classification, we define them as follows:
\begin{itemize}
    \item \textbf{Class-indicating Words} (CWs)
    \item \textbf{Fake Class-indicating Words} (FWs)
    \item \textbf{Class-irrelevant Words} (IWs)
\end{itemize}
\textit{CWs} are both semantically and statistically related to the class, which means they are good features of the corresponding class and can well describe the meaning of the class. These words are the overlap between the features of the dataset and the classes. \textit{FWs}, which are statistically related to the class in the dataset while not semantically related to the class, are usually misleading features brought by the limited or biased dataset. These words are quite troublesome since the classifier is hard to distinguish between real class-indicating features and fake class-indicating features. \textit{IWs} are those that have little correlation with the class, such as the stop-words, which usually don't have too much effect on the classification.

A text classifier with strong generalization ability should be able to recognize the class-indicating words, while not influenced by the fake class-indicating or class-irrelevant words. In practice, both CWs and FWs are likely to be learned as class-indicating features, thus the features from the FWs may mislead the prediction and hurt the generalization performance. Thereby, to train a better text classification model, especially when training samples are limited and even biased, we should try to improve the model's ability to identify the CWs and in the meanwhile reduce the influence of FWs to avoid the model from learning incorrect class features.

\section{Selectively Augment the Text for Classification}
The last section presents three kinds of keywords that play different roles in text classification tasks. These \textit{role keywords} provide us more insights about what should be learned by a text classifier to have good generalization ability. In this section, we first introduce how to effectively extract these \textit{role keywords} from the text. Based on these keywords, we propose \textit{Selective Text Augmentation} techniques to reinforce or diminish the words from the text according to their roles to generate better augmentation samples.

\begin{figure*}[t]
 \centering
\includegraphics[width=\textwidth]{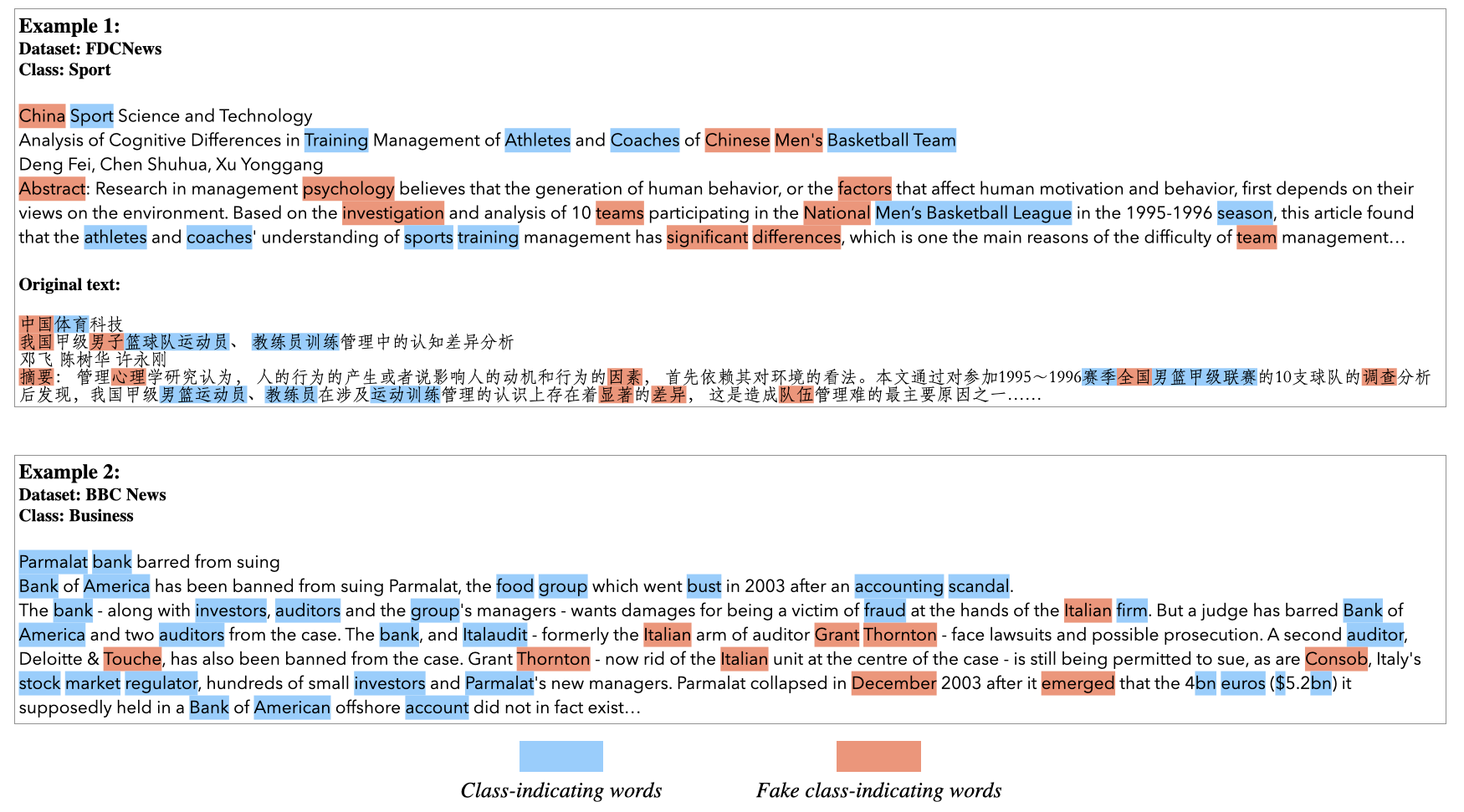} % Reduce the figure size so that it is slightly narrower than the column.
\caption{Examples of \textit{role keywords} extraction. }
\label{role_kws_extraction}
\end{figure*}

\subsection{Role keywords Extraction}
To extract the \textit{role keywords}, we should first find evaluation metrics for correlation and similarity level. For the measurement of correlation with the class, we employ weighted log-likelihood ratio (WLLR) to select the class-correlated words from the text sample. This is inspired by \cite{yu2016learning} where WLLR is used to find out the "pivot words" for sentiment analysis. The WLLR score is computed by:
\begin{align*}
    wllr(w,y) = p(w|y)log(\frac{p(w|y)}{p(w|\bar{y})})
\end{align*}
where $w$ is a word, $y$ is a certain class and $\bar{y}$ represents all the other classes in the classification dataset. $p(w|y)$ and $p(w|\bar{y})$ are the probabilities of observing $w$ in samples labeled with $y$ and with other labels respectively. We can use the frequency of a word occurring in the certain class to estimate the probability.

To measure the similarity between a word and the meaning of the class label, a straightforward way is to use word vectors pre-trained with skip-gram \cite{mikolov2013distributed} or Glove \cite{pennington2014glove}. We can compute the cosine similarity between a word and a class label to see their semantic distance:
\begin{align*}
    similarity(w,l) = \frac{v_w \cdot v_l}{\| v_w \|  \| v_l \|}
\end{align*}
where $l$ represents the label and $v_w$, $v_l$ are word vectors for the word $w$ and the label $l$. We can also use a description of the label to obtain $v_l$ by averaging the word vectors of each word in the description for better label representation. In our experiments, we find that simply using the word or phrase of the label itself is enough to measure the similarity between a word and the category.

We compute the WLLR scores and similarity scores of each word in a given sample, and set a ratio $\alpha$ to get the top words by each metric as the correlated words $W_{c}$ and similar words $W_{s}$. By combining these words, we can extract the \textit{role keywords} as follows:
\begin{align*}
    W_{cw} &= \{ w| w \in W_{c} \cap W_{s}\} \\
    W_{fw} &= \{ w| w \in W_{c} \cap \overline{W_{s}}\} \\
    W_{iw} &= \{ w| w \in \overline{W_{c}} \} 
\end{align*}
where $W_{cw}$, $W_{fw}$ and $W_{iw}$ are CWs, FWs and IWs respectively. In our experiments, we don't extract the IWs since we aren't using them for text augmentation. $\alpha$ is a hyper-parameter in \textit{role keywords} extraction which should be determined according to the dataset and the task. Larger $\alpha$ will increase the number of CWs and FWs while also increase the risk of incorrect extraction. Table \ref{ratio_for_extraction} is the statistics of the extraction precision and proportion of CWs by counting 100 samples randomly selected from FDUCNews dataset, which shows that with the increasing of $\alpha$, the precision of CW extraction is decreasing while the number of correct CWs is increasing. We recommend to choose $\alpha=0.2$ as a moderate setting to balance the quality and quantity of \textit{role keywords}.

\begin{table}[h]
\centering
\begin{tabular}{c|ccc}
\toprule[1pt]
    $\alpha$ & 0.1  & 0.2  & 0.3  \\ 
\midrule
\textit{precision} & 89\% & 86\% & 75\% \\
\textit{proportion}   & 3\%  & 8\%  & 12\% \\
\bottomrule
\end{tabular}
\caption{Effect of different $\alpha$ on the precision and proportion of CWs extraction. Proportion means the proportion of correctly extracted CWs in all the words of the sample. The choice of $\alpha$ is a trade-off between the quality and quantity of \textit{role keywords}.}
\label{ratio_for_extraction}
\end{table}

To illustrate the effectiveness of the above methods for \textit{role keywords} extraction, we exhibit the extraction results of some real training samples in our experiments in Figure \ref{role_kws_extraction}. Example 1 is a sample from the FDUCNews dataset with label "sport". We can see that words/phrases like "sports", "training", "athletes" and "Basketball Team" are extracted as CWs. The words such as "Chinese", "Abstract", "team" and "significant difference", are extracted as FWs. By checking the samples from FDUCNews dataset, we found that lots of documents from "sport" class are sourced from academic reports about Chinese sports, in which words like "Chinese", "Abstract" and "significant difference" are quite common. These words are semantically irrelevant to "sport" class, but are statistically correlated with "sport" class in this dataset. These words tend to be learned as misleading features towards "sport" class. Example 2 is a similar case where words like "\$", "euros", "firm", "market", "bank" and "investors" are screen out as CWs. Words like "Italian" and "December" are filtered out as FWs. 

Once the \textit{role keywords} are extracted, we can use them to guide our text augmentation.

\subsection{STA: Selective Text Augmentation}
In this paper, we mainly take advantage of text editing techniques including synonyms replacement, word order swap, word deletion and insertion for text augmentation and study how to upgrade them to a selective manner. Other augmentation methods like back-translation can also be transformed and will be explored in future.

Given a piece of text from the training set, \textbf{STA} will augment the text by the following selective text editing operations:
\begin{itemize}
    \item \textbf{Selective Replacement}: Select $n$ words from CWs and replace each of them with one of its synonyms or similar words. 
    \item \textbf{Selective Insertion}: We designed two kinds of insertion: \textbf{inner insertion} and \textbf{outer insertion}. The former method will select $n$ FWs from samples belonging to other classes into the given text. By doing this, we force the biased noisy features of one class to be uniform across all classes. The latter selects $n$ CWs and inserts their synonyms. 
    \item \textbf{Selective Swap}: Choose $n$ words from the CWs and swap them with another $n$ random words in the text.
    \item \textbf{Selective Deletion}: We also designed two kinds of deletion techniques. One we call it \textbf{noise deletion} and the other one \textbf{positive selection}. By noise deletion, we delete the FWs in the text to generate a new sample without potential misleading information. The positive selection on the other hand only selects the CWs from the text to form the new sentence of which all the words are highly related to the class.
\end{itemize}

\section{Experiments Setup}
\subsection{Datasets}
We conduct experiments on four benchmark text classification datasets, including two English datasets and two Chinese datasets: \\
$\bullet$ \textbf{NG}: The 20NG dataset\footnote{https://www.cs.umb.edu/~smimarog/textmining/datasets/} (bydata version) is an English news dataset that contains 18846 documents evenly categorized into 20 different categories.\\
$\bullet$ \textbf{BBC}: The BBC news categorization dataset\footnote{http://mlg.ucd.ie/datasets/bbc.html} is introduced in \cite{bbc} as a benchmark for document clustering and classification, which consists of 2225 documents from the BBC news website in five topical areas. \\
$\bullet$ \textbf{FD}: The FDCUNews dataset\footnote{http://www.nlpir.org} is a Chinese text classification dataset provided by Fudan University which contains 9833 Chinese news categorized into 20 different classes. Since some classes only contain a few samples, we select six classes (sport, computer, politics, economy, agriculture and environment) which  have enough samples to form our experimental dataset.\\
$\bullet$ \textbf{TH}: The THUCNews dataset\footnote{http://thuctc.thunlp.org} is a Chinese news classification dataset collected by Tsinghua University. We constructed a subset from it which contains 39000 news evenly split into 13 news categories.

\subsection{Baseline Methods}
We choose \textbf{EDA} introduced in \cite{wei2019eda} as the baseline which is composed of four text editing operations: 1) \textbf{random synonyms replacement} which randomly chooses $n$ words from the text to employ synonyms replacement, 2) \textbf{random insertion} which randomly selects $n$ words and inserts their synonyms, 3) \textbf{random swap} which randomly swaps the positions of $n$ pairs of words and 4) \textbf{random deletion} which randomly deletes $n$ words from the text. These techniques all showed obvious performance gain for text classification on small datasets in original paper.

\subsection{Settings}
During augmentation, we make use of the WordNet \cite{miller1995wordnet} to search the synonyms or similar words both in baseline methods and our methods in English datasets. For Chinese datasets, since there isn't a widely used Chinese synonym dictionary like WordNet, we use publicly available pre-trained skip-gram word embeddings to find the similar words as the synonyms. For the choice of $\alpha$ in role keywords extraction, we find that smaller $\alpha = \{0.1,0.2,0.3\}$ get similar results for subsequent selective augmentation, and $0.2$ is a moderate and robust setting across different datasets. Therefore we set $\alpha=0.2$ in all our experiments. We set the proportion of words changed during augmentation to be 10\%, which is a recommended setting in \cite{wei2019eda}. If the role keywords extracted from the text are less than 10\%, we than select random words from the text as supplements. For the choice of text classification model, we choose the popular TextCNN \cite{kim2014convolutional} which is widely used and proven to have good performance in text classification tasks to conduct our experiments. We use early-stopping with patience $p=3$ to choose the best model for evaluation.

\begin{table}[h]
\small
\centering
\begin{tabular}{cc|ccc}
            Dataset         &   $n$   & no aug. & EDA     & \textbf{STA}              \\
\toprule[1.5pt]
\multirow{3}{*}{\textbf{FD}}  & 500  & 50.34 & 71.60 & \textbf{74.71} \\
                     & 1000 & 65.73 & 81.68 & \textbf{83.55} \\
                     & 2000 & 80.88 & 85.43 & \textbf{86.94} \\
\midrule
\multirow{3}{*}{\textbf{TH}}  & 500  & 35.21 & 47.90 & \textbf{52.59} \\
                     & 1000 & 48.35 & 62.87 & \textbf{67.99} \\
                     & 2000 & 65.79 & 76.45 & \textbf{78.64} \\
\midrule
\multirow{3}{*}{\textbf{BBC}} & 500  & 61.97 & 73.14 & \textbf{75.28} \\
                     & 1000 & 78.31 & 87.06 & \textbf{88.31} \\
                     & 2000 & 87.89 & 91.92 & \textbf{92.91} \\
\midrule
\multirow{3}{*}{\textbf{NG}}  & 500  & 11.37 & 14.65 & \textbf{19.13} \\
                     & 1000 & 18.87 & 26.24 & \textbf{31.60} \\
                     & 2000 & 25.93 & 38.05 & \textbf{45.86} \\
\midrule
\multicolumn{2}{c}{\textit{Average}}     & 52.55 & 63.08 & \textbf{66.46}\\ 
\bottomrule[1.5pt]
\end{tabular}
\caption{Test accuracy (\%) on four benchmark datasets with different train size $n$. "no aug." means no augmentation is used. Both EDA and STA augment the training samples by 6 times, using all text editing operations.}
\label{Results_1}
\end{table}

\section{Experimental Results}
\subsection{STA Outperforms Non-selective Methods}
We evaluate STA on four benchmark datasets in a low resource setting with training set size $n=\{500,1000,2000\}$, compared with corresponding non-selective text augmentation method EDA. For each training sample, we use EDA or STA to generate 6 augmented samples (STA and EDA both have four kinds of operations, but insertion and deletion of STA both have two sub-operations, therefore insertion and deletion of EDA are both used twice for fair comparison). The results are shown in Table \ref{Results_1}.

From the results, we can see that STA significantly outperforms the non-selective counterpart EDA in all datasets, with an average accuracy improvement at about 3.4\%. Particularly, in NG dataset, which is a quite complicated text classification dataset with 20 classes and many sub-classes, STA outperforms EDA by 5.9\% on average. The improvements reveal that the augmented data generated by STA can help the model to achieve better generalization ability than traditional non-selective data augmentation methods.

\begin{table}[h]
\small
\centering
\begin{tabular}{l l| l l l} 
 \toprule[1.5pt]
 \multirow{1}{*} & \textbf{Methods} & $n$=500 & $n$=1000 &  $n$=2000 \\
 \midrule[1.0pt]
 & no aug. & 39.72 & 52.82 & 65.12 \\
 \midrule[1.0pt]
 \multirow{5}{*}{\textbf{EDA}} & replace\textsubscript{r} & 42.82 & 55.96 & 65.35 \\
 & insert\textsubscript{r}  & 45.63 & 57.83 & 69.45 \\
 & swap\textsubscript{r}  & 44.16 &56.76 &69.65 \\ 
 & delete\textsubscript{r}  &44.37	&58.63 &68.65 \\

 & EDA(mix)  & 51.82 & 64.46 & 72.96 \\
 \midrule[1.0pt]
  \multirow{7}{*}{\textbf{STA}} & replace\textsubscript{s}  & 40.13 & 54.87 & 65.99* \\
 & insert\textsubscript{s}(inner)  & 43.47 & 56.53 & 69.26 \\
 & insert\textsubscript{s}(outer) & 42.87 & 56.51 & 70.05* \\
 & swap\textsubscript{s}  & 44.43* & 57.65* & 70.31* \\
 & delete\textsubscript{s} (nd)  & 45.80* & 58.81* & 69.82* \\
 & delete\textsubscript{s} (ps)  & 51.39* & 64.50* & 72.88* \\

 & STA(mix)  & \textbf{55.43}* & \textbf{67.86}* & \textbf{76.09}* \\
 \bottomrule[1.5pt]
\end{tabular}
%}
\caption{Comparison of average test accuracy across all datasets using different text editing operations for augmentation. Subscript $r$ refres to "random (non-selective)" whereas subscript $s$ refers to "selective". "inner": inner insertion. "outer": outer insertion. "nd": noise deletion. "ps": positive selection. "mix" means using all sub-operations for augmentation. Note that for EDA(mix) insertion and deletion are both repeated twice for fair comparison with STA(mix). "*" means the selective method outperforms its non-selective counterpart.}
\label{Results_2}
\end{table}

\subsection{Comparison of Different Text Editing Operations}
To further investigate the effect of different text editing operations, we decompose EDA and STA to test the performance of each operation and report the results in Table \ref{Results_2}. There are some interesting observations: 1) Selective augmentation is not always better, especially for operations like synonyms replacement and synonyms insertion. Specifically, we can see from the results that when train size is only 500 or 1000, the performances of selective replacement and selective insertion are all slightly worse than that of non-selective manner. 2) Insertion and deletion are two most effective operations for \textit{non-selective} text editing augmentation while deletion and swap are most powerful for \textit{selective} augmentation in our experiments. 3) \textit{Positive selection} which is a sub-method of selective deletion is particularly effective. Specifically in our experiments, 
although the number of augmented samples of \textit{positive selection} is 6 times smaller than that of EDA, \textit{positive selection} can get comparable or even better results than EDA. 

Through our analysis, we make the following explanations for these observations: For synonyms replacement and insertion, though selectively augmenting those class-indicating words can bring more class-indicating features to the training set, the misleading fake class-indicating words still remain in the text which limits the performance of augmentation. Replacement or insertion at random however can bring in some noise to increase the model's robustness. Deleting those fake class-indicating words and class-irrelevant words from the training samples can help the classifiers to learn to distinguish between real and fake class-indicating features and pay more attention on the true features of the concerned categories. Therefore selective deletion can significantly improves the generalization performance of text classification models.

\begin{table}[h]
\small
\centering
\begin{tabular}{cc|ccc}
            Dataset         &   $n$   & no aug. & EDA     & \textbf{STA}              \\
\toprule[1.5pt]
\multirow{2}{*}{\textbf{TH}}  & 500  & 74.10 & 79.20 & \textbf{79.85} \\
                     & 2000 & 87.80 & 90.06 & \textbf{91.03} \\
\midrule
\multirow{2}{*}{\textbf{NG}}  & 500  & 40.09 & 47.80 & \textbf{49.54} \\
                     & 2000 & 70.64 & 73.50 & \textbf{74.48} \\
\bottomrule[1.5pt]
\end{tabular}
\caption{Test accuracy (\%) on experiments using BERT-based classifiers. The augmentation settings are consistent with the experiments in Table \ref{Results_1}.}
\label{Results_bert}
\end{table}

\subsection{Experiments Using Pre-trained Transformer-based Classifiers}
Previous studies show that these simple text-editing based augmentation methods might not yield substantial improvements when using pre-trained models\cite{wei2019eda}. Therefore we further conduct experiments on TH and NG datasets using pre-trained transformer-based classifiers to validate the performance of our proposed methods. Specifically, we use BERT-tiny\cite{bert-tiny} for NG dataset and Albert-tiny\cite{albert} for TH dataset. The results are shown in Table \ref{Results_bert}, which illustrate that our STA still outperforms EDA significantly in our experiments. However, the improvements are relatively smaller compared with the results using TextCNN in Table \ref{Results_1}. This may result from the strong capacity of pre-trained transformers which have already obtained great amount of knowledge about the target classification task. More experiments on other datasets and other experimental settings on transformer-based models will be conducted in future.

\begin{table}[h]
\small
\centering
\begin{tabular}{ll|ccc}
                     &   Methods   & FD(500) & TH(500)       \\
\toprule[1.5pt]
\multirow{2}{*}{}  & no aug.  & 91.21 &	74.10   \\
                    
\midrule
\multirow{4}{*}{\textit{TE-based}}  & EDA(d)  & 93.11 & 76.75   \\
                           & STA(ps) & 93.79 & 77.78  \\
                           & EDA(mix)  & 93.32 & 79.20 \\
                           & STA(mix) & \textbf{93.99} &	\textbf{79.85}  \\
\midrule
\multirow{2}{*}{\textit{LM-based}}  & BT  & 93.65 & 77.57   \\
                           & CA  & 93.86 & 76.52  \\
\bottomrule[1.5pt]
\end{tabular}
\caption{Comparison between TE-based and LM-based augmentation methods. BT: Back Translation. CA: Contextual Augmentation. "mix" means using all TE operations. "EDA(d)" refers to "random deletion" while "STA(ps)" refers to "positive selection" which is a strong sub-method of STA. We use BERT-based classifiers here as in Section 5.3.}
\label{Results_lm}
\end{table}

\section{Related Work}
Previous text augmentation methods can be divided into two types. One is to apply all kinds of text editing techniques including replacement \cite{sr2011,sr2015a,sr2015b}, insertion \cite{wei2019eda},  position swap \cite{wei2019eda} and deletion \cite{16-xie2017data}. \cite{wei2019eda} proposed EDA package as an integration of these text editing methods for text augmentation. Text editing methods for augmentation are easy to implement and have been proven to provide obvious improvements for smaller dataset. The other kind is using generative models such as pre-trained language models for back-translation \cite{9-xie2019unsupervised,back_translation1,back_translation2}, synonyms replacement \cite{8-jiao2019tinybert,lm_for_sr,kobayashi2018contextual} and sentences synthesizing \cite{anaby2020not}.  Although these type of augmentation methods are powerful, they are relatively inconvenient due to the high implementation cost of complex models. In addition to these methods, recently more and more creative and complex methods have been designed. For instance, \cite{mixup} uses the idea of mixup on NLP tasks, \cite{RL} leverages reinforcement learning for text augmentation. 

\subsection{Comparison with LM-based augmentation methods}
This work mainly focuses on the text-editing(TE)-based augmentation methods, including replacement, insertion, swap and deletion, and explore how to improve their performances by a selective manner. Thus, we mainly reported the comparison with EDA, which consists of four different TE operations and is regarded as a good representative of traditional TE-based augmentation methods. 

As a reference, we also compared our TE-based STA with two famous but much more complicated language model(LM)-based methods, Back Translation (\textbf{BT})\cite{back_translation1, back_translation2} and Contextual Augmentation (\textbf{CA})\cite{kobayashi2018contextual}. In our experiments, BT uses a ZH-EN NMT model for back-translating, CA uses a pre-trained BERT model for synonyms replacement with replacement probability at 0.1. Both BT and CA augment the text once. Since the augmentation procedures of BT and CA are quite time-consuming, we only conducted experiments on FD and TH datasets with train size at 500. The results are reported in Table \ref{Results_lm}. From the results, we can see that BT and CA are both strong augmentation techniques. In FD dataset, augmenting by BT or CA only once can beat the performance of EDA(mix) which augments six times. However, a single augmentation by "positive selection", a sub-method of STA, outperforms both BT and CA in these two datasets, and the numbers can be further improved by using STA(mix) with other operations together. What's more, since STA doesn't need to use large pre-trained language models for generating augmented samples, we can use STA to generate more training samples than LM-based methods in the same amount of time.

Since we only conducted experiments to compare these methods on two small datasets with the default configuration of BT and CA, we are not meant to show that our STA is definitely better than these more complicated LM-based methods. Actually, we believe different methods have different suitable application scenarios, and we encourage readers to try different augmentation methods when time and resource are sufficient.

\section{Conclusion \& Future Work}
In this paper, we first systematically summary three kinds of \textit{role keywords} each of which has unique effect on the classification. We illustrate how the gap between the features of the dataset and the features of the classes can influence the classifier's generalization ability, and how the extracted \textit{role keywords} can guide us to train a better model. We then propose methods to effectively extract these \textit{role keywords} and present STA to selectively generate training samples which can help the model to learn more useful information from the limited data while avoid being mislead by the noise and bias from the dataset. Extensive experiments reveal the effectiveness of STA over traditional non-selective augmentation methods.

Though STA have displayed significant advantages over traditional non-selective text augmentation methods, it still has room for improvement. First, STA depends on the extraction of \textit{role keywords}. Although we provide methods to extract them, the boundaries between different types are not necessary optimal. There exists better approaches to extract these \textit{role keywords} and should be studied in future. Second, our proposed selective augmentation techniques for different text editing operations can be further improved, especially for replacement and insertion. More complicated ways of selective augmentation can be designed to improve the performance.

In fact, STA is only one application of the \textit{role keywords} to show its effectiveness. The idea of \textit{role keywords} provides us insights of what have been learned by a classifier while also tells us what should be learned to train a good classifier, which is not only limited to text classification. Our methods have the potential to translate to other domains apart from the text. For example, the idea of \textit{role keywords} in this work might inspire research in \textit{role superpixels} in computer vision tasks for image augmentation, since different parts of images also have different roles in classification. Our future work includes the following directions: 1) Applying STA on domain adaptation problems, where the cross-domain prediction suffers from the mismatching of the feature spaces of different domains, which may be alleviated by selective augmentation. 2) Generalizing the idea of selective augmentation to computer vision for image augmentation.

\newpage
\bibliography{anthology,custom}
\bibliographystyle{acl_natbib}

\end{document}